\documentclass[runningheads]{llncs}
\usepackage[T1]{fontenc}
\usepackage{graphicx}
\usepackage{hyperref}

\hypersetup{
    colorlinks=true,
    linkcolor=blue,
    citecolor=blue,
    urlcolor=blue
}
\usepackage{booktabs}
\usepackage{adjustbox}
\usepackage{multirow}
\usepackage{amsmath}
\usepackage{amssymb}
\usepackage{bm}
\usepackage{xcolor}
\usepackage{subcaption}
\usepackage[misc]{ifsym}

\begin{document}
\title{Exploring the Transfer Learning Capabilities of CLIP in Domain Generalization for Diabetic Retinopathy}
\titlerunning{CLIP for DRDG} 
%

\author{Sanoojan Baliah \textsuperscript{\Letter} \orcidID{0009-0001-0326-2968}  \and
Fadillah A. Maani \orcidID{0000-0001-5927-7782} \and
Santosh Sanjeev\orcidID{0000-0003-3664-3844} \and
Muhammad Haris Khan \orcidID{0000-0001-9746-276X}}
\authorrunning{S. Baliah et al.}
%
\institute{Mohamed bin Zayed University of Artificial Intelligence, Abu Dhabi, UAE
\email{\{sanoojan.baliah, fadillah.maani, santosh.sanjeev, muhammad.haris\}@mbzuai.ac.ae}
}
\maketitle              
\begin{abstract}

Diabetic Retinopathy (DR), a leading cause of vision impairment, requires early detection and treatment. Developing robust AI models for DR classification holds substantial potential, but a key challenge is ensuring their generalization in unfamiliar domains with varying data distributions. To address this, our paper investigates cross-domain generalization, also known as domain generalization (DG), within the context of DR classification. DG, a challenging problem in the medical domain, is complicated by the difficulty of gathering labeled data across different domains, such as patient demographics and disease stages. Some recent studies have shown the effectiveness of using CLIP to handle the DG problem in natural images. In this study, we investigate CLIP's transfer learning capabilities and its potential for cross-domain generalization in diabetic retinopathy (DR) classification. We carry out comprehensive experiments to assess the efficacy and potential of CLIP in addressing DG for DR classification. Further, we introduce a multi-modal fine-tuning strategy named Context Optimization with Learnable Visual Tokens (CoOpLVT), which enhances context optimization by conditioning on visual features. Our findings demonstrate that the proposed method increases the F1-score by 1.8\% over the baseline, thus underlining its promise for effective DG in DR classification. Our code is publicly available at {\url{https://github.com/Sanoojan/CLIP-DRDG}}.

\end{abstract}


\section{Introduction}
Deep learning (DL) has become the standard for Computer Vision tasks, achieving state-of-the-art performance. However, DL models often suffer significant performance degradation when training and testing distributions differ \cite{ben2006analysis,bendiff}, including in medical imaging, where diverse patient characteristics and scanners present challenges in deploying DL solutions globally. Additionally, data safety and privacy concerns limit access to data in medical centers. In medical imaging, it is intuitive to consider different medical centers as different domains. Thus, the research on domain generalization (DG) \cite{muandet2013domain,ghifary2015domain,zhou2020learning,khan2021mode} can be a solution to widely mitigate major health problems. 


Diabetic Retinopathy (DR) is a major global health concern, leading to blindness as a complication of diabetes \cite{kempen2004prevalence}. Relying solely on clinicians for early prevention is impractical due to the complexity and variability in diagnosing DR. To address this, researchers have attempted to develop automatic systems for accurate DR diagnosis from fundus images \cite{he2016deep,dosovitskiy2020image,touvron2021training}. Typically, DR stage diagnosis involves identifying specific lesions, which requires expert annotation for each image. To tackle this issue, several works have focused on deep learning-based DR classification \cite{Asiri2019,bodapati2021composite,Wu2020}, categorizing DR into five distinct groups \cite{kempen2004prevalence}: no DR, mild, moderate, severe, and proliferative.


ImageNet pre-trained models benefit DR classification \cite{Asiri2019} indicating that models trained on natural images can still benefit the downstream DR classification task although the domain is very different. The CLIP model, trained on 400 million image-text pairs using contrastive learning, exhibits excellent multi-modality and zero-shot performance on diverse natural image datasets \cite{DBLP:journals/corr/abs-2109-01903,CLIPradford2021learning}. CLIP's success extends to tasks beyond classification, including object detection and video classification, owing to its joint vision and text contrastive learning approach. CLIP's success in zero-shot learning and transfer learning for natural tasks has inspired researchers to explore its potential in medical domains \cite{clipsegmentation,pubclip,medclip}. However, little to no work has been done on evaluating CLIP's performance for DR classification due to the lack of meaningful medical reports for each image sample. This study aims to investigate the potential of adopting CLIP methodology for DR in the domain generalization experiment setting.

Motivated by CLIP's generalizable capability in natural images, this work investigates CLIP performance for DR grading in the DG experiment setting by adopting CLIP with various strategies. Our key contributions are,
\textbf{1.}
 We conduct an extensive analysis of CLIP multi-modalities for DR grading in the DG setting with various strategies for adapting CLIP architecture.
    \textbf{2.}
 We investigate various text modalities in the CLIP architecture. We show that combining the CLIP visual encoder and BioBERT \cite{biomedicalbert} text encoder pre-trained on biomedical domain corpora results in a better performance. \textbf{3.}
 We propose  \textbf{Context Optimization with Learnable Visual Tokens} (\textbf{CoOpLVT}), a CLIP-based architecture that leverages image-conditioned prompt tuning with visual backbone fine-tuning. We also demonstrate how our proposed method performs better in terms of F1 score.

\section{Related Work}

\noindent\textbf{Domain generalization (DG):} The earliest study on DG, known as empirical risk minimization (ERM) \cite{vapnik1999nature}, aimed to minimize errors across source domains. Variants like multi-task autoencoders \cite{ghifary2015domain}, maximum mean discrepancy (MMD) constraints \cite{muandet2013domain}, and contrastive learning approaches \cite{motiian2017unified,dou2019domain,Kim_2021_ICCV} have been proposed. DomainBed \cite{Gulrajani2021InSO} introduced a fair evaluation protocol for DG, showing competitive performance for ERM. SWAD \cite{cha2021swad} proposed stochastic weight averaging to achieve flatter minima for DG.

\noindent \textbf{Domain generalization in Medical Imaging:} In medical imaging analysis, diverse data distributions from different sources often lead to reduced model performance in new environments. However, few studies address the DG problem in medical imaging. One approach, presented in \cite{li2022domain}, utilized task-specific augmentations to enhance data diversity and employed episodic learning. Another work from \cite{li2020domain} captured the shareable information by leveraging variational encoding to learn a representative feature space through linear-dependency regularization. Recently, for diabetic retinopathy (DR) classification, the DRGen method \cite{atwany2022drgen} combined the SWAD approach \cite{cha2021swad} for achieving flatter minima and the Fishr technique \cite{rame2022fishr} for regularization to match gradient variations across domains.

\noindent \textbf{Large Scale models in Medical Domain:}
BioBERT\cite{biomedicalbert} and ClinicalBERT\cite{huang2020clinicalbert} are two examples of domain-specific language models that have been pre-trained on large-scale biomedical and clinical text datasets, respectively. The pre-training process involves fine-tuning a preexisting BERT model on a large biomedical or clinical text corpus, allowing the model to learn domain-specific features and terminology.  

\noindent \textbf{CLIP and its adoption in DG:}
CLIP \cite{CLIPradford2021learning} has achieved significant success in image-text pre-training and supports various downstream tasks in computer vision and natural language processing. It has been applied to DG problems in image classification, demonstrated in \cite{bose2023stylip}, \cite{zhang2022domain}, and \cite{niu2023domainunified}. In the medical domain, CLIP has been explored for segmentation \cite{clipsegmentation} and Medical Visual Question Answering (MedVQA) \cite{pubclip}. However, adapting vision-text pre-training to the medical domain is challenging due to limited datasets and subtle differences within medical domains. \cite{convirt}, \cite{huang2021gloria}, and \cite{medclip} address these challenges with contrastive learning strategies, but there are no works that fully explore CLIP's domain generalizability in the medical domain.

\noindent \textbf{Prompt Engineering: }
Recent works like \cite{coop} and \cite{zhou2022conditional} propose methods for generating effective prompts to enhance vision-language models, demonstrating improved performance on benchmark datasets. \cite{zhou2022conditional} introduces dynamic prompts that adapt to each instance, making them less sensitive to class shift and achieving better results in visual question-answering and image captioning tasks compared to fixed and learned prompts.

\section{Methodology}

%


The DG problem requires learning from multiple source domains or seldom a single source, and evaluating on an unseen target domain. Since target and source data distribution are different, the goal is to learn cross-domain generalizable features.
Specifically, we characterize each domain $d$ by $\mathcal{D}^{d}=\{(\mathbf{x}_i^{d},y_i^d)\}_{d=1}^{n}$ that consists of samples drawn from
i.i.d. with a probability $\mathcal{P}(\mathcal{X}^d,\mathcal{Y}^d)$,
where $\mathbf{x}\in\mathcal{X}$ is an image with $C \times H \times W$ shape and $y \in \mathcal{Y} = \{1,2,..,K\}$ is an associated class label.
In multi-source domain generalization,
 a model is trained using data from multiple domains, i.e., the number of training domains ($d_{tr}$) $> 1$.
We sample $b$ instances from every training domain for each step, making the total batch size of $B=b \times d_{tr}$ \cite{Gulrajani2021InSO}.
We target the task of classification, and the model can be represented as $\mathcal{F}=w\circ f_v$, which maps the input images to the target label space, where  $f_v: \mathbf{x} \rightarrow h$ is the visual encoder and $w:h \rightarrow y$ is the classifier.
Then, we evaluate the model generalization on data sampled from $\mathcal{P}(\mathcal{X}^{d_{te}},\mathcal{Y}^{d_{te}})$, where $d_{te}$ represents the target domain.



\subsection{CLIP Adoption Techniques} 
\textbf{Empirical Risk Minimization (ERM):} ERM \cite{vapnik1999nature} uses the Cross Entropy (CE) loss over all samples in mini-batch to optimize a predictor. Under the DG settings, ERM simply aggregates data from all source domains 
and searches for the optimal predictor $\mathcal{F}$ by minimizing the empirical risk
$\frac{1}{B} \sum_{i=1}^B \mathcal{L}_{\mathrm{CE}}(\mathcal{F}(\mathbf{x}_i),y_i)$. A recent work \cite{Gulrajani2021InSO} employed ERM for DG problem and showed competitive performance in DG datasets under a fair evaluation. Owing to the 
increasing adoption of Vision Transformers (ViT) for various Computer Vision tasks,
there is a growing interest in benchmarking their DG performance. This naturally suggests the usage of ViT as the feature encoder $f$ in the ERM pipeline, namely ERM-ViT.

\begin{figure}[t]
    \centering
    \includegraphics[width=\textwidth]{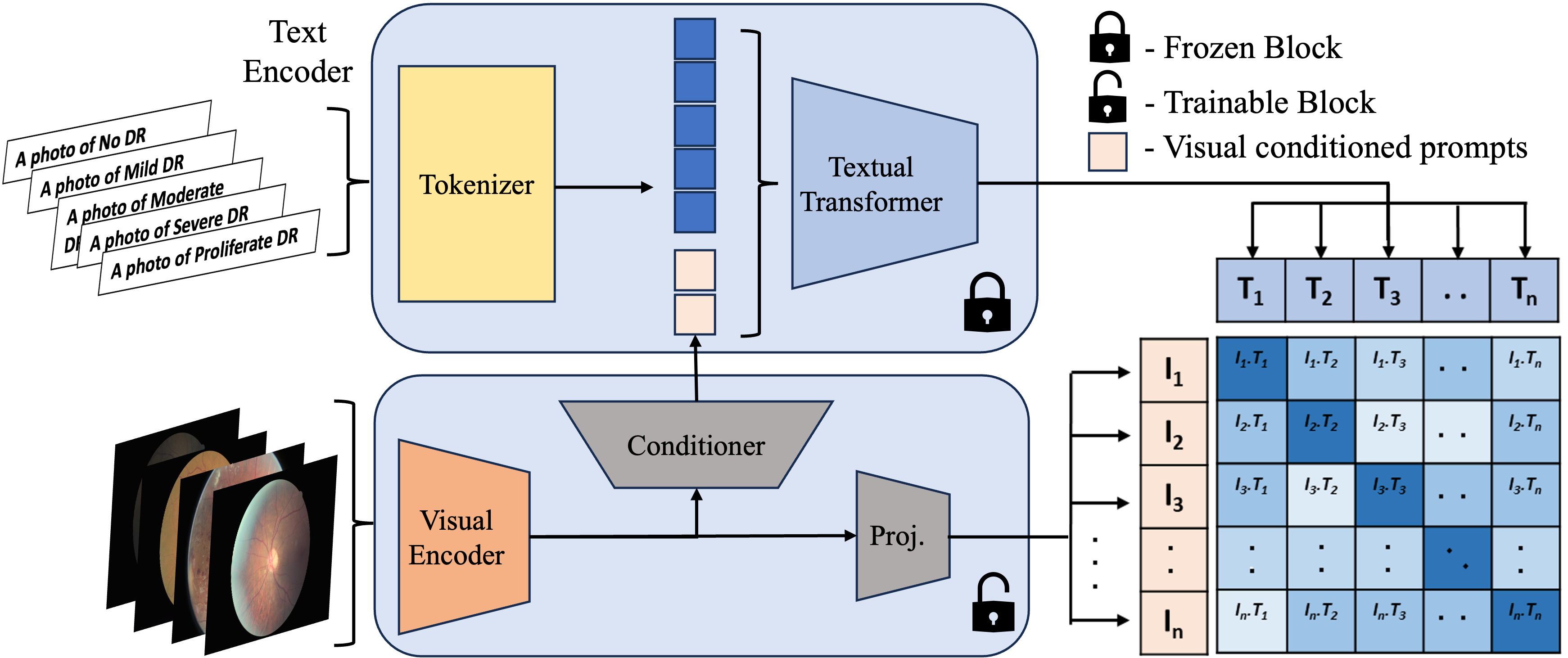}
    \caption{Overall architecture of our proposed Context Optimization with Learnable Visual Tokens (\textbf{CoOpLVT}).}

    \label{fig_cooplvt}
\end{figure}
\noindent\textbf{Linear Probing:}
 Given the prominent generalizability of CLIP pre-trained models 
 on fine-tuning \cite{kumar2022finetuning} and zero-shot \cite{DBLP:journals/corr/abs-2109-01903} settings, a rather simple alternative of adopting CLIP to medical imaging can be performed by retraining the classifier $w$ while keeping the CLIP backbone $f_v$ frozen.
\noindent\textbf{Naive Multi-modal Fine-tuning:}
We train the CLIP's visual encoder on the similarities between visual features $V$ and text features $T$ by applying cross-entropy loss, i.e. $-\frac{1}{B}\sum_{j=1}^B y_jlog(\sigma(\langle V,T \rangle))$ where $\langle , \rangle$ symbolizes dot product. Note that it is a simple fine-tuning strategy under the zero-shot setting. More specifically, given $K$ number of classes, we generate $K$ text prompts (denoted as $S$) based on the class names following the CLIP's text template, e.g. ``a photo of a \{class name\}". These text prompts are then encoded into textual features $T=f_t(S) \in \mathbb{R}^{S \times C_f}$, where $f_t$ is the text encoder and $C_f$ is the output feature dimension. For each optimization step, we sample  $B$ number of images (minibatch size) and obtain $B$ visual features using CLIP visual encoder as $V=f_v(\bm{x}) \in \mathbb{R}^{B \times C_f}$. For architecture diagrams of ERM-ViT, Linear Probing and Naive CLIP refer to the supplementary material.

Further, we choose to only optimize CLIP's visual encoder while keeping the text encoder fixed. Training the full network might be crucial as medical domain labels are quite different from the natural image cases, however, this creates a challenge of high GPU memory consumption. To this end, we propose our fine-tuning strategy next, which considers both modalities while training, unlike fixed textual feature representation in the above approaches.

\subsection{Context Optimization with Learnable Visual Tokens}

Apart from the analysis of the CLIP with general fine-tuning approaches for DR, inspired from CoCoOp \cite{zhou2022conditional}, we introduce a multi-modal fine-tuning strategy, namely
\textbf{CoOpLVT},
which trains the visual model and a conditioner to interact with textual transformer.  To formally introduce the approach, let us decompose the visual feature $V$ as $V=f_v(\bm{x})=p \circ f_I(\bm{x})$, where $p$ is a linear projector often used in multi-modal models to bring both modalities' features to the same dimension, and $f_I$ is the immediate higher dimension ($d^I$) feature from the CLIP visual encoder. Also, we denote the conditioner $\mathcal{G}:\mathbb{R}^{d^I} \rightarrow \mathbb{R}^{d'}$ an MLP network.

Since our motivation is to incorporate context-aware text prompts to interact on the textual encoder, we concatenate additional $N_p$ tokens to the original text tokens created from the CLIP's text generation template mentioned in the Naive CLIP approach. As shown in Fig.~\ref{fig_cooplvt}, to generate these context-aware tokens, we introduce the learnable $\mathcal{G}$, where $d'=N_p \times d^T$. Here, $d^T$ is the feature dimension of textual tokens. Thus, the text feature for an image $j$ can be generated as,
\begin{equation}
    T^j=f_t([S[y_j],\mathcal{G}(f_I(x^j))])
\end{equation}
where $S[y_j]$ is the corresponding template prompt tokens for the image $j$ and $f_t$ is a transformer-based text encoder that enables us to concatenate any number of context-aware tokens which can interact with the frozen tokens in self-attention.
Unlike CoCoOp, we train the visual encoder as the CLIP has not seen any DR data. 
With the use of context-aware text tokens, we create different text features for all the images in the batch, i.e. all the images have different visual and textual feature pairs.  This further allows us to formulate the loss as in the original CLIP contrastive loss \cite{radford2019language}, even on the downstream classification task with limited class settings. Thus, our proposed loss can be formulated as, 
\begin{equation}
    \mathcal{L}_{\text{contrastive}}(V, T)=\underbrace{-\frac{1}{B} \sum_{j=1}^B j \log \left(\sigma\left\langle V^j, T\right\rangle\right)}_{\mathcal{L}_{\text{text}}} \underbrace{-\frac{1}{B} \sum_{j=1}^B j \log \left(\sigma\left\langle V, T^j\right\rangle\right)}_{\mathcal{L}_{\text{visual}}}
\end{equation}
where $\mathcal{L}_{visual}$, and $\mathcal{L}_{text}$ are the logits when we consider the similarity matrix in visual and text direction respectively.




\section{Experiments}
\begin{table}[h]
\centering
\caption {Multi-Source DG Results(F1-score). Bold numbers mean best and underlined are the second best, and () denotes the standard deviation. CLIP-V, CLIP-VB, B(p), and B(s) denote CLIP visual-encoder, CLIP visual-encoder with BioBert text-encoder, BioBert pre-trained on PubMed, and BioBert pre-trained on SQuAD, respectively.}

\adjustbox{max width=\textwidth}{%
\begin{tabular}{ll|rrrrr}

\toprule
\multirow{2}{*}{\textbf{Strategy}} & \multirow{2}{*}{\textbf{Algorithm}} &  \multicolumn{5}{c}{\textbf{F1-score}} \\ \cmidrule{3-7}
 &  & \multicolumn{1}{c}{\textbf{APTOS}} & \multicolumn{1}{c}{\textbf{EyePACS}} & \multicolumn{1}{c}{\textbf{Messidor}} & \multicolumn{1}{c}{\textbf{Messidor2}} & \multicolumn{1}{c}{\textbf{Avg}}\\
\midrule

\multirow{2}{*}{ERM}        & ResNet50   & \underline{28.6 (0.8)} & 29.3 (0.4) & 45.8 (0.9) & 51.3 (0.7) & 38.8 (0.5)\\
                            & ViT        & 24.0 (1.6) & 30.9 (1.0) & 46.6 (0.3) & 53.4 0.6) & 38.7 (0.3) \\

\midrule
\multirow{3}{*}{Zero-shot}  & CLIP       & 3.4 (0.1)  & 4.4 (0.0)  & 4.0 (0.1)  & 2.2 (0.1)  & 3.5 (0.0) \\
                            & CLIP-VB(p) & 14.6 (0.0) & 18.4 (0.0) & 15.9 (0.1) & 12.7 (0.1) & 15.4 (0.0)\\
                            & CLIP-VB(s) & 11.6 (0.1) & 17.9 (0.0) & 12.7 (0.1) & 14.7 (0.1) & 14.2 (0.0)\\

\midrule
Linear probing              & CLIP-V & 13.0 (0.2) & 14.0 (2.5) & 12.4 (0.0) & 14.2 (0.6) & 13.4 (0.5)\\
\midrule

\multirow{2}{*}{Naive CLIP} & CLIP       & 26.0 (1.0) & 30.7 (1.0) & \textbf{47.7 (0.4)} & \underline{53.2 (0.5)} & 39.4 (0.7) \\
                            & CLIP-VB(p) & 26.5 (0.6) & \underline{31.6 (0.7)} & \underline{46.5 (0.5)} & \textbf{53.5 (0.3)} & \underline{39.5 (0.3)} \\

\midrule

\textbf{CoOpLVT} (ours)           & CLIP       & \textbf{31.9 (2.6)} & \textbf{32.2 (0.3)} & 46.2 (0.7) & 51.6 (0.3) & \textbf{40.5 (0.5)} \\
\bottomrule

\end{tabular}
}
\label{tablef1}
\end{table}
\noindent\textbf{Datasets:} Following the work of \cite{atwany2022drgen}, we validate the effectiveness of our approach on four datasets, EyePACs \cite{EyePACS}, APTOS \cite{aptos}, Messidor and Messidor-2 \cite{ImageAnalStereol1155} which serve as 4 domains in the DG setting. EyePACs is the largest dataset with 88702 fundus images, while APTOS, Messidor, and Messidor-2 have 3657, 1200, and 1744 fundus images, respectively which are classified into 5 classes: no DR, mild, moderate, severe, and proliferative DR \cite{kempen2004prevalence}. However, Messidor does not contain any proliferative samples. Following DRGen \cite{atwany2022drgen}, we use images of size 224 $\times$ 224, with the random flip, random grayscaling, color jittering, random rotation, translation, and Gaussian blur augmentations.

\noindent\textbf{Implementation and training/testing details:} For a fair comparison, we follow the default domainbed \cite{Gulrajani2021InSO} settings throughout all the experiments. Specifically, we experiment with the 80:20 training, validation splits. We use the batch size ($b$) of 32 per domain making the total batch size 96, the learning rate of $5e-05$ for Imagenet trained models and the $5e-06$ for CLIP trained models, and no weight decay. We use the AdamW optimizer with the default Pytorch params. 
 We train and validate the models on the source domains while the unseen target domain is used for testing only. We report accuracy by repeating the same experiment 3 times with different trial seeds (on two V100 GPUs, implemented in Pytorch). We strictly follow, not to access the test domain as a validation set (i.e. non-oracle model selection). 

\begin{table*}
\centering
\caption {Single Source domain generalization results (f1 score) .}
\scalebox{0.97}{
\tabcolsep=0.08cm
\begin{tabular}{lcccccccc}
\toprule

\textbf{Algorithm}  &  \textbf{Train Dataset} &\textbf{APTOS}         & \textbf{EyePACS}     & \textbf{Messidor}    
& \textbf{Messidor2}  & \textbf{Avg}         \\
\midrule

Naive CLIP  & APTOS      &   -       & 26.2     & 13.4     & 22.8    & 20.7       \\ 
CoOpLVT (Ours)  & APTOS      &   -       & \textbf{28.1}  & \textbf{14.5}  & \textbf{25.5 } & \textbf{22.7}     \\ 
\midrule

Naive CLIP  & EyePACS      &   \textbf{42.9 }     & -      & 30.1     & 44.9    & 39.3      \\ 
CoOpLVT (Ours)  & EyePACS       &  40.7   & -       & \textbf{34.9 }    &  \textbf{48.4} & \textbf{41.3}        \\ 
\midrule

Naive CLIP  & Messidor      &  20.7 &  20.3  &  -   &  37.3   & 26.1      \\ 
CoOpLVT (Ours)  & Messidor   &   \textbf{23.6}  & \textbf{22.4}    & -       &  \textbf{38.9}    &  \textbf{28.3}     \\ 
\midrule

Naive CLIP  & Messidor2      &   33.9 &   29.3 &    \textbf{47.9}   &   -  &   37.0       \\ 
CoOpLVT (Ours)  & Messidor2   & \textbf{38.4}      &  \textbf{29.7}   & 47.1       & -       & \textbf{38.4}      \\ 
\bottomrule
\end{tabular}}
\label{singleDomain}
\end{table*}






\noindent\textbf{Metrics}: Two widely-used image classification metrics are used to evaluate our experiments: accuracy and F1-score. We include accuracy as it has been used in previous research on DG for DR (See supplementary), allowing us to benchmark our experiments with other works. However, given the highly imbalanced class distribution in DR, it is intuitive to choose the F1-score as the primary metric since it considers both precision and recall. Thus, in this work, we aim to explore and improve CLIP performance on DR with DG setting based on the F1-score.

\noindent\textbf{Experiments}.We conduct experiments to explore various strategies for adopting the CLIP model for tackling the DR problem in the DG setting. (1) We start with utilizing a pre-trained CLIP visual encoder as a DR classification model trained with the ERM algorithm. (2) Inspired by the decent performance of CLIP zero-shot in natural images, we inspect CLIP zero-shot performance on DR with two sets of prompts and several text encoders. (3) Linear probing is also investigated with CLIP visual encoder, as it is capable of extracting distinguishing features for general classification problems. (4) We experiment with naive multi-modal fine-tuning. (5) We evaluate our proposed \textbf{CoOpLVT} through various ablation studies. Table \ref{tablef1} provides the results of the main experiments, and we further discuss the experiments in Section \ref{sec:results_and_discussion}.

\section{Results \& Discussion} \label{sec:results_and_discussion}


Initially, we compare the effectiveness of CLIP pre-trained models with the Imagenet pre-trained models  and also fine-tune the full network using the ERM-ViT approach (see Table \ref{tablef1}). We explore the zero-shot capabilities of CLIP and also conduct other experiments changing the text-encoders. The zero-shot performance of CLIP is very poor as CLIP was trained only on the natural domain and has not seen much of medical-related data. We replace the CLIP's text encoder with BioBERT pre-trained on PubMed and SQuAD. We observe a boost in the performance of the models in both accuracy and F1-score. This is due to the fact that BioBERT was trained on large scale medical data and has a very good zero-shot performance on medical-related tasks.
\begin{table}
\centering
\caption{Ablation study on zero-shot CLIP performance. \textbf{Prompt I}: \textit{"a photo of a \{c\}"} where c belongs to [\textit{No DR}, \textit{mild DR}, \textit{moderate DR}, \textit{severe DR}, \textit{proliferative DR}]. \textbf{Prompt II}: similar to Prompt I, with \textit{"a photo of a \{c\}"} where c belongs to [\textit{No Diabetic Retinopathy}, \textit{mild Diabetic Retinopathy}, \textit{moderate Diabetic Retinopathy}, \textit{severe Diabetic Retinopathy}, \textit{proliferative Diabetic Retinopathy}].}

\adjustbox{max width=\textwidth}{%
\begin{tabular}{l|rrrrrr}

\toprule

\textbf{Model} & \multicolumn{2}{c}{CLIP} & \multicolumn{2}{c}{CLIP-VB(p)} & \multicolumn{2}{c}{CLIP-VB(s)} \\
\textbf{Prompt} & \multicolumn{1}{c}{I} & \multicolumn{1}{c}{II} & \multicolumn{1}{c}{I} & \multicolumn{1}{c}{II} & \multicolumn{1}{c}{I} & \multicolumn{1}{c}{II} \\

\cline{2-7}

\textbf{F1-Score} & 1.2 (0.0) & 3.5 (0.0) & \textbf{15.4 (0.0)} & 8.1 (0.0) & 14.2 (0.0) & 7.9 (0.1) \\
\textbf{Acc} & 3.1 (0.1) & 4.9 (0.1) & 44.3 (0.1) & 16.4 (0.0) & \textbf{47.9 (0.1)} & 19.6 (0.1) \\

\bottomrule

\end{tabular}
}
\label{ablation_zero_shot}
\end{table}

\noindent \textbf{Multi source results:} Linear probing for CLIP gives 56.4\% average accuracy and 13.4\% average F1-score, suggesting the CLIP pre-trained weights, without fine-tuning, cannot be applied to DR as the CLIP is trained mostly on natural images. Further, when train with the Naive-CLIP approach by replacing the text-encoder with BioBERT\cite{biomedicalbert} pre-trained on PubMed we obtain 39.4\%  and 39.5\% average F1-score respectively. In Naive-CLIP approach the prompts are fixed and it is a very challenging task to design the prompts as they are task-specific and require domain expertise. To overcome this problem, we design CoOpLVT which generates the best performance with an average F1-score of 40.5\%.

We also conduct ablation studies to understand the effect of using different prompt designs (see Table \ref{ablation_zero_shot}). From the results we observe that slight changes in the prompts can have a huge effect on the F1-score and accuracy. We further conduct different experiments by freezing and fine-tuning the visual encoder, varying the number of MLP layers as well as the number of induced visual tokens (can be seen in supplementary material). We observed that increasing the number of MLP layers helps in improving the performance whereas increasing the number of induced tokens causes a drop in the performance.

\noindent \textbf{Single source DG results:} We show the comparison of single source DG results between Naive CLIP and CoOpLVT (Top two performing strategies in multi-source DG). CoOpLVT consistently provides a better F1 score than Naive CLIP. \textbf{Attention maps:} Further, the analysis of attention maps (see supplementary) reveals that the CLIP model can find DR lesions in the eye in most cases.




\section{Conclusion}

In this work, we explore the effectiveness of the CLIP pre-trained model and its generalizability through various sets of experiments. To our knowledge, we are the first to investigate the DG performance of the CLIP model in medical imaging, especially in DR. We investigate and analyze the performance of zero-shot as well as fine-tuned settings. We investigate the effectiveness of the CLIP pre-trained model and its generalizability through various transfer learning techniques. In addition, we propose a multi-modal fine-tuning strategy named CoOpLVT to suit DR data. With extensive experiments, we showed the capabilities of CLIP for domain generalization in DR and demonstrated that our proposed approach results in a better F1-score by 1.8\% compared to the baseline performance.
{\small
\bibliographystyle{splncs04}
\bibliography{egbib}
}






\end{document}


\title{Supplementary material for: Exploring the Transfer Learning Capabilities of CLIP in Domain Generalization for Diabetic Retinopathy}
%
%
\author{}
\authorrunning{}
\institute{}

%
\maketitle              

\begin{figure}[t]
    \centering
    \includegraphics[width=\textwidth]{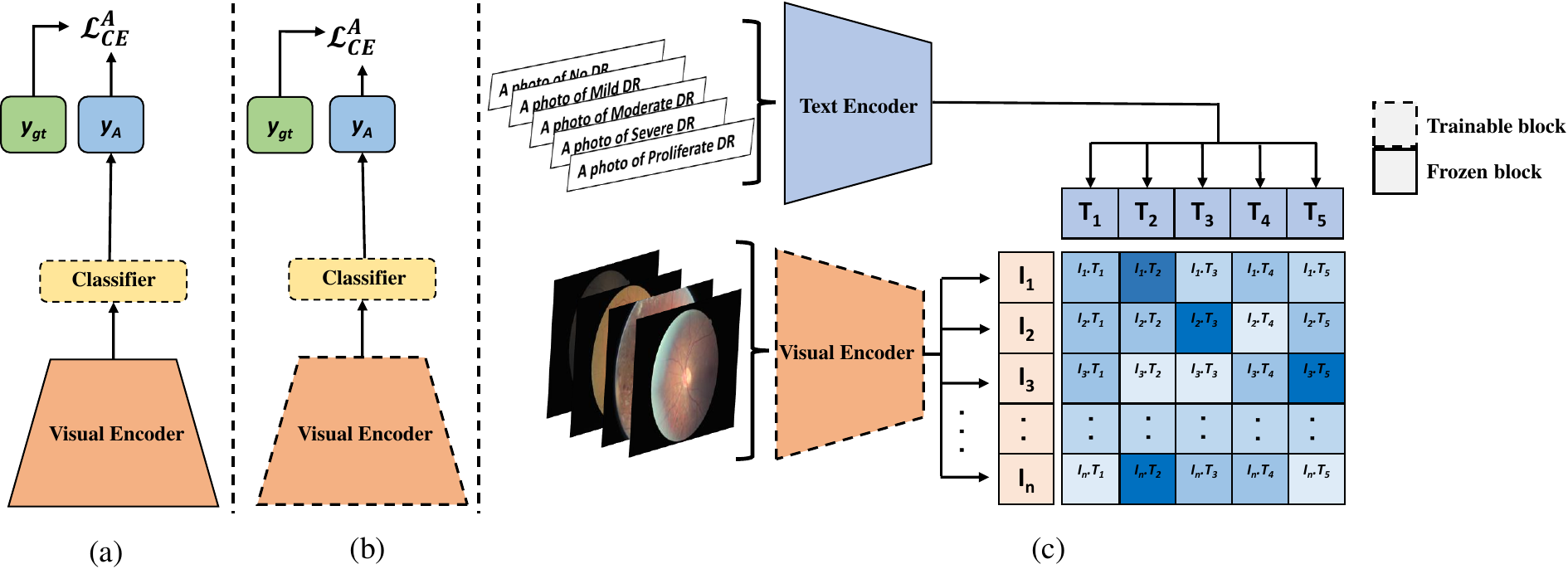}
    \caption{ This diagram illustrates, the methods we use to adopt the CLIP pre-trained model, in DR. (a) shows the linear probing, (b) shows the ERM-ViT where the entire model is fine tuned and (c) shows the naive CLIP approach in which we use the matrix-softmax loss over the similarity scores between the text and visual features. 
    }
    \label{fig_main}
\end{figure}

\begin{figure}[htb!]
    \centering
    \includegraphics[width=\textwidth]{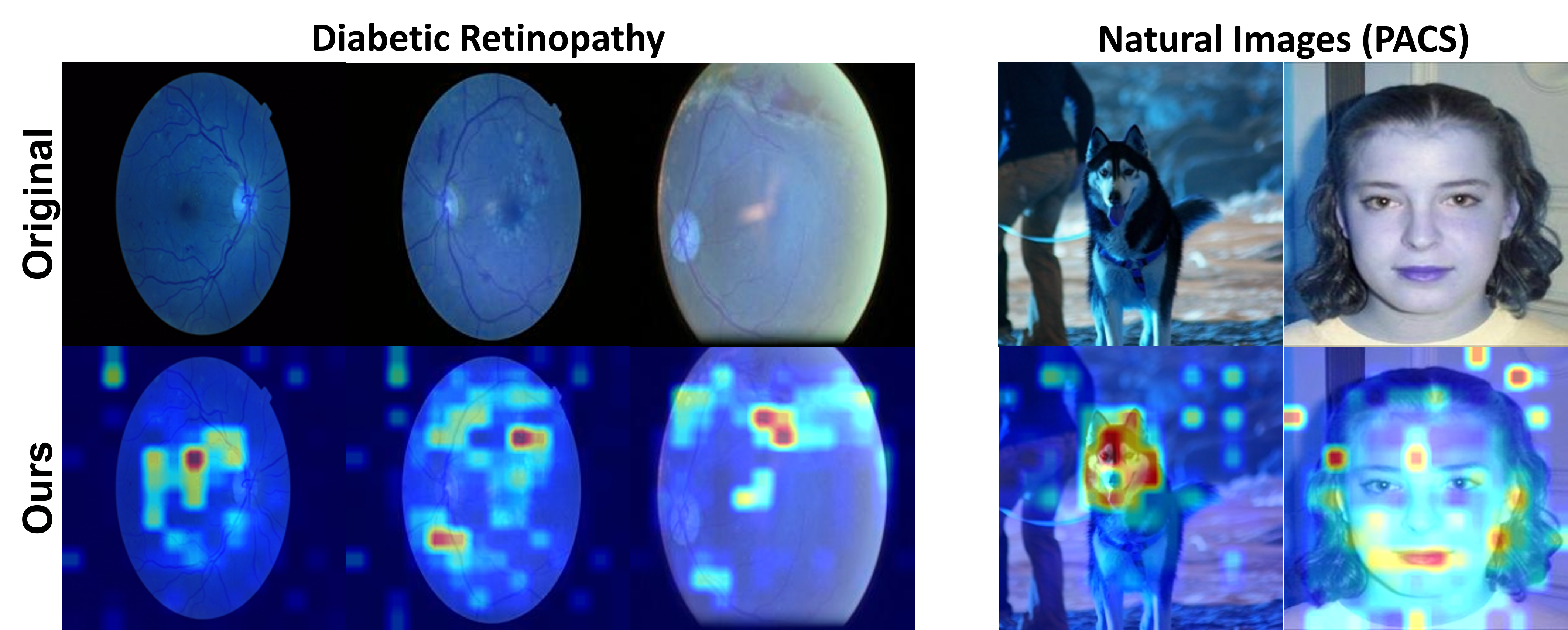}
    \caption{ Attention visualization of some proliferate DR and severe DR cases with our approach at the last self-attention layer. Even though the CLIP model can find DR lesions in the eye in most cases, it also attends to some unwanted positions. We verify with the PACS-picture domain to check whether it is the case with natural images as well. As a general observation, CLIP gives better attention to some classes like animals and birds, and worse attention to classes like faces, and buildings regardless of a very high accuracy (99\%).
    }
    \label{fig_atten}
\end{figure}

\begin{table}[h]
\centering
\caption {Multi-Source DG Results(Accuracy). Bold numbers mean best and underlined are the second best, and () denotes the standard deviation. CLIP-V, CLIP-VB, B(p), and B(s) denote CLIP visual-encoder, CLIP visual-encoder with BioBert text-encoder, BioBert pre-trained on PubMed, and BioBert pre-trained on SQuAD, respectively.}

\adjustbox{max width=\textwidth}{%
\begin{tabular}{ll|rrrrrr}

\toprule
\multirow{2}{*}{\textbf{Strategy}} & \multirow{2}{*}{\textbf{Algorithm}} &  \multicolumn{5}{c}{\textbf{Accuracy}} \\
 &  & \multicolumn{1}{c}{\textbf{APTOS}} & \multicolumn{1}{c}{\textbf{EyePACS}} & \multicolumn{1}{c}{\textbf{Messidor}} & \multicolumn{1}{c}{\textbf{Messidor2}} & \multicolumn{1}{c}{\textbf{Avg}}\\
\midrule
DRGen              & Resnet50 & \underline{51.3(3.0)}      & 72.6(0.8)      & 60.8(1.1)       & 66.0(1.5)       & 62.6(-) \\
\midrule

\multirow{2}{*}{ERM}        & ResNet50 & 50.0(2.0)      & 68.9(1.2)       & 63.5(1.2)       & 67.6(3.1)      & 62.5(0.2)  \\
                            & ViT        & 46.3(1.9) & 68.5(2.1) & 64.0(0.9) & 69.1(1.4) & 62.0 (1.1) \\

\midrule
\multirow{3}{*}{Zero-shot}  & CLIP       & 8.2(0.2)  & 4.2(0.0)  & 4.6(0.1)  & 2.6 (0.1) & 4.9 (0.1)  \\
                            & CLIP-VB(p) & 47.0(0.0) & 59.5(0.0) & 38.9(0.2) & 31.8(0.3) & 44.3 (0.1) \\
                            & CLIP-VB(s) & 26.3(0.1) & 64.0(0.0) & 43.9(0.3) & 57.4(0.3) & 47.9 (0.1) \\

\midrule
Linear probing              & CLIP-V & 49.3(0.0) & \textbf{73.5(0.0)} & 44.8(0.2) & 58.1(0.3) & 56.4 (0.1)\\
\midrule

\multirow{2}{*}{Naive CLIP} & CLIP       & \textbf{51.4(1.2)} & 71.0(2.1) & \textbf{66.9(0.4)} & 66.6(1.1) & \underline{64.0 (0.7)} \\
                            & CLIP-VB(p) & 47.3(1.0) & \underline{73.0(0.3)} & \underline{66.0(0.4)} & \underline{70.1(1.2)} & \textbf{64.1 (0.5)} \\

\midrule

\textbf{CoOpLVT} (ours)           & CLIP       & 46.2(0.7) & 65.9(2.0) & 65.5(0.4) & \textbf{70.6(0.6)} & 62.1 (1.4) \\
\bottomrule

\end{tabular}
}
\label{tableacc}
\end{table}

\begin{table}[h]
\centering
\caption{Ablation study on the first proposed method, comprising of (a) trainable parameters, (b) the number of MLP layers in the conditioner, and (c) the number of prompts.}

\adjustbox{max width=\textwidth}{%
    \begin{subtable}{0.47\textwidth}
        \centering
        \begin{tabular}{lrr}
        \textbf{Trainable} & \textbf{F1-Score} & \textbf{Acc} \\
        \hline
        Conditioner & 1.2 (0.0) & 3.1 (0.1) \\
        Conditioner & \multirow{2}{*}{\textbf{40.0 (0.5)}} & \multirow{2}{*}{\textbf{61.6 (0.8)}} \\
        + CLIP-V & & \\
       \end{tabular}
       \caption{Freezing vs. fine-tuning the visual-encoder.}
       \label{ablation_clip_method_i_a}
    \end{subtable}

    \begin{subtable}{0.47\textwidth}
        \centering
        \begin{tabular}{lrr}
        \textbf{\# MLP} & \textbf{F1-Score} & \textbf{Acc} \\
        \hline
        1-layer & 40.0 (0.5) & 61.6 (0.8) \\
        3-layers & \textbf{40.5 (0.5)} & \textbf{62.1 (1.4)} \\
        \\
       \end{tabular}
       \caption{The conditioner MLP design.}
       \label{ablation_clip_method_i_b}
    \end{subtable}
    
    \begin{subtable}{0.47\textwidth}
        \centering
        \begin{tabular}{crr}
        \textbf{\# Prompts} & \textbf{F1-Score} & \textbf{Acc} \\
        \hline
        4 & \textbf{40.5 (0.5)} & \textbf{62.1 (1.4)} \\
        8 & 38.4 (0.3)  & 60.3 (0.8) \\
        \\
       \end{tabular}
       \caption{The number of induced visual tokens.}
       \label{ablation_clip_method_i_c}
    \end{subtable}
}

\label{ablation_clip_method_i}
\end{table}






